\documentclass[runningheads]{llncs}

\usepackage[mobile]{eccv}

\usepackage{eccvabbrv}
\usepackage{multirow}
\usepackage{multibib}
\usepackage{algorithm} 
\usepackage{algpseudocode} 
\usepackage{graphicx}
\usepackage{booktabs}
\usepackage{marvosym}
\usepackage[accsupp]{axessibility}  
\newcites{appendix}{Appendix References}

\usepackage[pagebackref,breaklinks,colorlinks,citecolor=eccvblue]{hyperref}

\usepackage{orcidlink}
\makeatletter
\def\blfootnote{\xdef\@thefnmark{}\@footnotetext}
\makeatother
\begin{document}

\title{Supervised Fine-tuning \textit{in turn} Improves Visual Foundation Models} 

\titlerunning{ViSFT}

\author{Xiaohu Jiang\inst{1,2} \and
Yixiao Ge\inst{2,3}\textsuperscript{\Letter} \and
Yuying Ge\inst{3} \and \\
Dachuan Shi\inst{1} \and
Chun Yuan\inst{1}\textsuperscript{\Letter} \and
Ying Shan\inst{2,3}}

\authorrunning{X.~Jiang et al.}

\institute{Shenzhen International Graduate School, Tsinghua University \and
ARC Lab, Tencent PCG \and
Tencent AI Lab \\
\email{\{jiangxh21,\,sdc21\}@mails.tsinghua.edu.cn,}\\
\email{\{yixiaoge,\,yuyingge,\,yingsshan\}@tencent.com,}\\
\email{yuanc@sz.tsinghua.edu.cn}\\
\url{https://github.com/TencentARC/ViSFT}} 

\maketitle
\blfootnote{\noindent $^{*}$This work was done when Xiaohu Jiang was interning at ARC Lab, Tencent PCG. \textsuperscript{\Letter}Corresponding author.}

\begin{abstract}
  Image-text training like CLIP has dominated the pretraining of vision foundation models in recent years. Subsequent efforts have been made to introduce region-level visual learning into CLIP's pretraining but face scalability challenges due to the lack of large-scale region-level datasets. Drawing inspiration from supervised fine-tuning (SFT) in natural language processing such as instruction tuning, we explore the potential of fine-grained SFT in enhancing the generation of vision foundation models after their pretraining. Thus a two-stage method ViSFT (Vision SFT) is proposed to unleash the fine-grained knowledge of vision foundation models. In ViSFT, the vision foundation model is enhanced by performing visual joint learning on some in-domain tasks and then tested on out-of-domain benchmarks. With updating using ViSFT on 8 V100 GPUs in less than 2 days, a vision transformer with over 4.4B parameters shows improvements across various out-of-domain benchmarks including vision and vision-linguistic scenarios.
\end{abstract}

\section{Introduction}
\label{sec:intro}

Training of vision foundation models has witnessed significant progress in recent years~\cite{dosovitskiy2020image,touvron2021training,caron2021emerging,oquab2023dinov2,radford2021learning,jia2021scaling, he2022masked,sun2023eva}. Among these developments, the image-text representation learning, exemplified by models such as CLIP~\cite{radford2021learning}, has become the mainstream approach for training vision foundation models, achieving state-of-the-art performance across various vision and vision-linguistic tasks. Furthermore, efforts like GLIP~\cite{li2022grounded} and RegionCLIP~\cite{zhong2022regionclip} aim to extend CLIP's capabilities by learning region-level visual representations during pretraining, thereby facilitating fine-grained downstream vision tasks. However, these efforts face scalability challenges due to the lack of large-scale region-level datasets.

\begin{figure}[t]
\centering
\includegraphics[width=.4\textwidth]{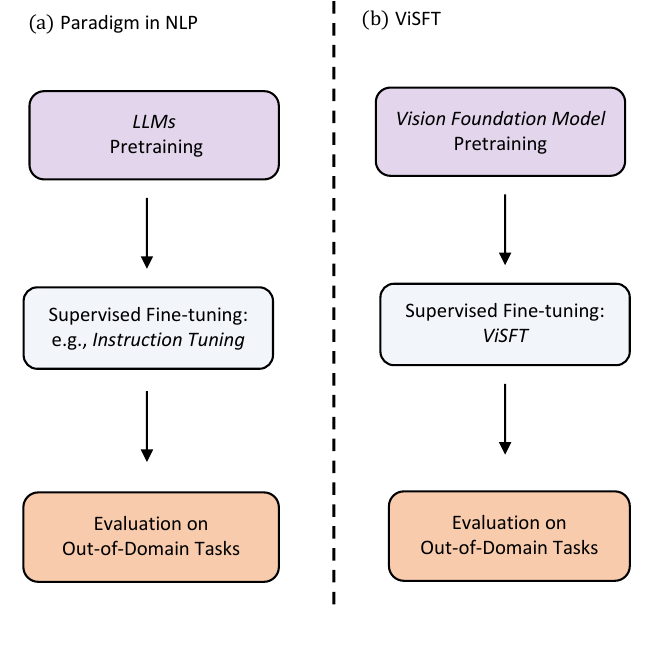}

\caption{Drawing inspiration from the training paradigm in NLP, we perform ViSFT on vision foundation models after their pretraining and subsequently evaluate them on out-of-domain tasks.}
\label{fig:inspiration}
\vspace{-1.5em}
\end{figure}

In the realm of natural language processing, the aforementioned challenge is addressed by employing supervised fine-tuning (SFT) following the pretraining of large language models, such as through instruction tuning~\cite{wei2021finetuned,sanh2021multitask,longpre2023flan,zhou2023lima,ivison2022hint}. By generating detailed task descriptions as instructions, the model undergoes SFT to understand and follow the instructions. Drawing inspiration from the NLP SFT, we investigate the potential of implementing pure \textbf{Vi}sion \textbf{SFT} (which we term \textbf{ViSFT}) to enhance the generalization capabilities of vision foundation models as shown in Figure~\ref{fig:inspiration}.

Our findings suggest that the representation and generalization of the vision transformer within a CLIP model can indeed be improved following ViSFT. In essence, ViSFT is able to unleash fine-grained details within the visual transformer that may have been overlooked during image-text pretraining. We speculate that this method assists the vision transformer in identifying a more optimal subspace.

In ViSFT, we incorporate the visual transformer as the backbone network connected to the heads of various in-domain vision tasks for joint learning. We opt for object-level tasks on COCO~\cite{lin2014microsoft}, including detection, segmentation, and captioning. Researchers commonly employ different LoRA~\cite{hu2021lora} weights to retain task-specific knowledge. Similarly, we use LoRA weights to preserve the unleashed information. Another benefit of LoRA tuning is its lightweight nature, which lowers training costs. 

ViSFT distinguishes itself from previous multi-task training approaches~\cite{ye2022taskprompter,hu2021unit,ye2022inverted,li2023uni,caruana1997multitask,crawshaw2020multi}, which fine-tune on in-domain task training splits and then maximize performance on validation splits. Our goal is to obtain fine-grained information through the joint learning of in-domain tasks, ultimately develop a vision transformer backbone with enhanced representation. To assess the generation capabilities of the improved vision model, a suitable choice involves evaluating its performance on out-of-domain benchmarks.


Another challenge lies in ensuring that knowledge learned from in-domain tasks can be effectively transferred to the vision transformer backbone, rather than being trapped in task heads. To address this, we divide ViSFT into two stages. In the first stage, we train the corresponding in-domain task heads independently while keeping the vision transformer backbone frozen. In the second stage, we introduce LoRA parameters to the vision transformer backbone and freeze the task heads, enabling the knowledge to be transferred exclusively to the LoRA parameters. The first stage of our approach allows us to obtain compatible in-domain task heads. This alleviates the necessity of devising intricate mechanisms to resolve domain conflicts, a common issue encountered in previous multi-task approaches.

Our experiments demonstrate that by undergoing ViSFT updating on $8$ V$100$-SXM$2$-$32$GB GPUs in less than 2 days, a CLIP vision transformer with a model size exceeding $4.4$B exhibits improvements across 5 different benchmarks, including vision and vision-linguistic scenarios (despite not performing SFT on the CLIP's text encoder). Our contributions can be summarized as follows:
\begin{enumerate}
    \item  We showcase the potential of fine-grained supervised fine-tuning (SFT) in enhancing the generalization capabilities of vision foundation models.
    \item A two-stage ViSFT process is proposed to effectively unleash the fine-grained knowledge of vision foundation models.
    \item The performance of visual foundation models has exhibited enhancements across a wide range of benchmarks in both visual and vision-linguistic scenarios, achieved through lightweight training.
\end{enumerate}

\section{Related work}

\subsection{Pretraining of Vision Foundation Model}
Pretraining of vision foundation models has experienced considerable progress in recent years. Following the introduction of the Vanilla Vision Transformer (ViT)~\cite{dosovitskiy2020image}, numerous pretraining paradigms have been explored for vision transformers, including supervised pretraining on large-scale image datasets~\cite{deng2009imagenet,sun2017revisiting}, self-supervised learning strategies~\cite{caron2021emerging,oquab2023dinov2}, masked image modeling techniques~\cite{he2022masked,peng2022beit}, and more. Notably, image-text pretraining methods~\cite{radford2021learning,jia2021scaling,yu2022coca} such as CLIP have emerged as the predominant approach for training foundational vision models. This method leverages extensive image-text data to pretrain models, aiming to learn the correspondence between images and text.

Moreover, efforts like GLIP~\cite{li2022grounded} and RegionCLIP~\cite{zhong2022regionclip} intend to introduce region-level visual representation learning into CLIP's pretraining process, thereby enhancing the performance of fine-grained downstream vision tasks. However, these endeavors encounter challenges in scaling up the model size due to the scarcity of large-scale region-level detection and grounding data. As a result, CLIP remains the prevailing paradigm in visual representation learning, supported by extensive image-text datasets. 

Recent EVA-CLIP series~\cite{fang2023eva,fang2023eva2,sun2023eva} achieve state-of-the-art performance on several zero-shot benchmarks. EVA first performs masked image modeling on scratch-based vision transformers to reconstruct the features of a CLIP's vision encoder. Then, the vision encoder of CLIP is replaced with the trained vision transformers for image-text pretraining. EVA successfully scales the vision transformer to over $4.4$ billion parameters. While BLIP-2~\cite{li2023blip} employs a bridge model (q-former) to integrate EVA-CLIP-G with large language models (LLMs), achieving state-of-the-art performance on various visual-linguistic benchmarks. Our ViSFT has explored the potential of fine-grained supervised fine-tuning in enhancing the generalization capabilities of both EVA-CLIP and BLIP-2.

\subsection{Visual-Linguistic Instruction Tuning}
Visual-linguistic instruction tuning represents a simple yet effective supervised fine-tuning (SFT) strategy for enhancing the generalizability of foundational models. Notably, natural language processing (NLP) instruction tuning~\cite{wei2021finetuned,sanh2021multitask,longpre2023flan,zhou2023lima,ivison2022hint} has achieved promising results in zero-shot learning by utilizing a small number of examples and a set of natural language instructions to guide the model in learning new tasks. There are generally two methods for constructing instruction datasets: data integration from annotated natural language datasets~\cite{longpre2023flan,sanh2021multitask} and generating outputs using LLMs~\cite{xue2023instruction,wang2022self}. Based on the collected IT dataset, a pre-trained model can be directly fine-tuned in a fully-supervised manner. Among these techniques, HINT~\cite{ivison2022hint} adopts a hypernetwork to convert instructions into adapter and prefix parameters, which is akin to how ViSFT stores fine-grained information in LoRA parameters.

Besides text-only domains, instruction tuning has been applied in multimodal domains~\cite{xu2022multiinstruct,liu2023visual,brooks2023instructpix2pix,gong2023multimodal,zhao2023svit}. MUL-TIINSTRUCT~\cite{xu2022multiinstruct} is a multimodal instruction tuning dataset comprising $62$ diverse tasks in a unified seq-to-seq format. LLaVA ($13$B)~\cite{liu2023visual} is a large multimodal model developed by connecting the visual encoder of CLIP ($400$M)~\cite{radford2021learning} with the language decoder LLaMA ($7$B)~\cite{touvron2023llama}. GPT-$4$ is employed to convert image-text pairs into an appropriate instruction-following format for LLaVA's dataset. While the above studies have achieved success in text-only and multimodal domains, the vision-only domain SFT has not yet been extensively explored.

\subsection{Multi-Task Training}
Multi-task training employs foundation models as the backbone, coupled with multiple task-specific heads. Typically, multi-task training involves fine-tuning the backbone and task-specific heads concurrently on downstream tasks' training splits and maximizing performance on validation splits, which are in-domain.

There has been extensive development in multi-task training across vision~\cite{he2017mask,zamir2018taskonomy,strezoski2019many,standley2020tasks,zamir2020robust}, language~\cite{sogaard2016deep,hashimoto2016joint,liu2017adversarial,sanh2019hierarchical,liu2019multi}, and multimodal domains~\cite{kaiser2017one,kiela2017learning,pramanik2019omninet}. Recent efforts aim to perform multi-task training using a single, generic model~\cite{kaiser2017one,li2023uni,ye2022inverted,ye2022taskprompter,zhu2022unimoe}. However, such attempts often face challenges due to task and domain conflicts, leading to the development of domain alignment methods and mechanisms to mitigate task conflicts.

ViSFT departs from traditional multi-task training approaches by obtaining fine-grained information through joint learning of in-domain tasks while evaluating performance on out-of-domain tasks. Additionally, rather than tuning LoRA and task heads simultaneously, ViSFT is divided into two stages. In the first stage, ViSFT can obtain compatible in-domain task heads independently, which alleviates the necessity to adopt task alignment mechanisms when the number of in-domain tasks increases. This makes the approach more flexible and easier to implement. To evaluate the enhancements of the vision model's generation capabilities, ViSFT focuses on the improvements made in out-of-domain tasks.

\section{Method}

\subsection{Tasks and Datasets}

\begin{figure*}[t]
\centering
\includegraphics[width=1.\textwidth]{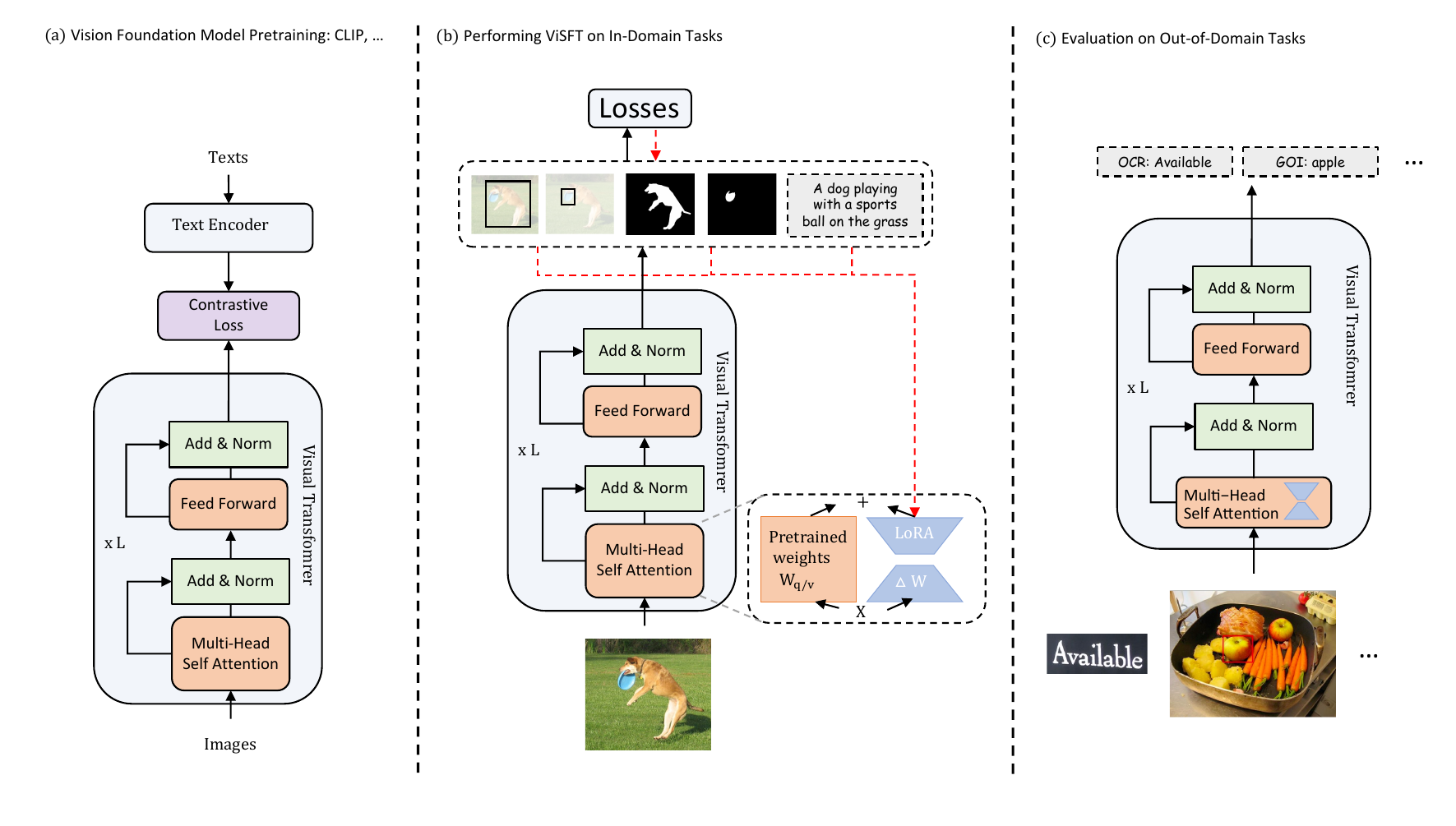}
\caption{An overview of our proposed method is as follows: (a) First, a vision foundation model is pretrained such as CLIP-ViT. (b) Next, we execute ViSFT to update the LoRA weights and retain the fine-grained information through joint learning of in-domain tasks. (c) Finally, in conjunction with the updated LoRA weights, evaluations on multiple out-of-domain tasks exhibit considerable enhancement. ``OCR" refers to the optical character recognition task, while ``GOI" denotes the grounded object identification task.}
\label{fig:overview}
\vspace{-0.6em}
\end{figure*}

To ensure that ViSFT remains both simple and fine-grained while eliminating the need to create new datasets, we opted to train our model using the COCO~\cite{lin2014microsoft} dataset. This dataset provides a diverse range of annotations for each image, including bounding boxes, instance-specific segmentation masks, natural language descriptions, and panoptic segmentation masks (a combination of instance and semantic segmentation). Additionally, 250k-person instances are annotated with keypoints. As depicted in Table~\ref{tab:annos}, these annotations facilitate the implementation of fine-grained learning.

Following the ablation studies in Sec~\ref{ablation}, we ultimately selected object detection, instance segmentation, and image captioning as the in-domain tasks. Moreover, tasks on COCO offer a variety of off-the-shelf task heads, obviating the need to develop new task heads.

\setlength{\tabcolsep}{3pt}
\renewcommand{\arraystretch}{1.1}
\begin{table}
    \caption{An overview of task categories and annotations in COCO, along with their associated task heads for implementation. Annotations excluded from our proposed solution are denoted in \textcolor{gray}{Gray}.}
  \label{tab:annos}
  \centering
  \resizebox{0.85\linewidth}{!}{
  \begin{tabular}{l|c|c}
    \toprule
    Tasks & Annotations  & Heads \\
    \midrule
    Object Detection & bounding boxes with 80 object categories & Detr \\
    Instance Segmentation & per-instance segmentation masks & Mask2former \\
    Image Captioning &  natural language descriptions of the images  & LSTM\\
    \textcolor{gray}{Panoptic Segmentation}& \textcolor{gray}{full scene segmentation with thing and stuff}   & \textcolor{gray}{Mask2former}    \\
    \textcolor{gray}{Pose Estimation} & \textcolor{gray}{person instances labeled with keypoints} & \textcolor{gray}{VitPose}\\
    \bottomrule
  \end{tabular}}
\end{table}

\subsection{Model Details}

In this section, we outline the process of conducting ViSFT on the vision foundation model as illustrated in Figure~\ref{fig:overview}. The entire model training procedure is divided into two stages. During the first stage, we employ the pre-trained vision transformer from an EVA-CLIP model to serve as the backbone network and freeze it. Detection, segmentation, and caption heads are then independently connected for fine-tuning. This step aims to obtain task heads that are compatible with the vision transformer features. In the second stage, the vision transformer is augmented with LoRA weights, and all task heads are connected for fine-tuning. Aside from the added LoRA weights, other modules will remain frozen. This approach ensures that fine-grained information obtained through joint learning is directed towards the LoRA parameters.

\vspace{0.5em}\noindent \textbf{EVA Vision Transformer.}
We select the vision transformer from EVA-CLIP~\cite{sun2023eva} as the vision foundation model, given its state-of-the-art performance and the architecture that is basically consistent with the vanilla ViT~\cite{dosovitskiy2020image}. As demonstrated in Table~\ref{tab:vit_size}, we conducted experiments using two model sizes: EVA-ViT-G and EVA-ViT-E.

\setlength{\tabcolsep}{3pt}
\renewcommand{\arraystretch}{1.1}
\begin{table}
\caption{Details of EVA-ViT model variants employed in our experiments: EVA-ViT-G and EVA-ViT-E, both with over 1 Billion parameters, are derived from EVA-CLIP-G and EVA-CLIP-E models, respectively.}
  \label{tab:vit_size}

  \centering
  \resizebox{0.8\linewidth}{!}{
  \begin{tabular}{l|c|c|c|c|c|c}
    \toprule
    Model & Layers  & Hidden size & Patch size & MLP size & Heads & Params \\
    \midrule
    EVA-ViT-G~\cite{sun2023eva} & 40 & 1408 & 14 & 6144 & 16 & 1B \\
    EVA-ViT-E~\cite{sun2023eva} & 64 & 1792 & 14 & 15360 & 16 & 4.4B \\
    \bottomrule
  \end{tabular}}
  
\end{table}

\noindent \textbf{LoRA Update Matrices.}
For a pre-trained weight matrix $W_{q/v} \in \mathbb{R}^{d\times k}$ within the query and value embedding layers of EVA-ViT, we impose a constraint on their updates by introducing a low-rank decomposition: $W_{q/v} + \Delta W = W_{q/v} + BA$, where $B \in \mathbb{R}^{d \times r}$ and $A \in \mathbb{R}^{r \times k}$, and rank $r < min(d,k)$. During the second stage of training, the weight matrices $W_{q}$ and $W_{v}$ are frozen, preventing them from receiving gradient updates, while $A$ and $B$ contain trainable parameters. For $h_{q/v} = W_{q/v}x$, the forward pass yields:
\begin{equation}
    h_{q/v} = W_{q/v}x + \Delta Wx = W_{q/v}x + BAx.
\end{equation}

\vspace{0.5em}\noindent \textbf{Detection Head.}
Among the available detection heads, Detr~\cite{carion2020end} is the first to incorporate transformers, which simplifies the detection head design, eliminates the need for intricate post-processing techniques such as non-maximum suppression, and supports single-scale feature input from vision transformers. 


Detr generates a fixed number of learnable query embeddings, which serve as input to the image decoder. These queries interact with one another via self-attention and interact with flattened image features through cross-attention layers. Subsequently, MLP and linear heads are employed for bounding box and label prediction, respectively. Finally, a bi-partite matching mechanism is used to assign predictions to ground truth boxes.

\vspace{0.5em} \noindent \textbf{Segmentation Head.} We utilize Mask2former~\cite{cheng2022masked} as the segmentation head. As a unified framework for segmentation tasks, Mask2former is capable of handling both instance segmentation and panoptic segmentation tasks, thereby providing convenience for experimenting with various segmentation annotations. To facilitate the use of vision transformers as the backbone, we have modified the input feature levels of Mask2former to $1$.

Mask2former also generates a fixed number of query embeddings. The segmentation mask representations are derived from the dot product between the decoder's final-layer hidden state of the $i$-th embedding and a per-pixel feature map:
\begin{equation}
    q_i^{\text{mask}} = \text{Upsample}\Big(\text{MLP}(q_i) \odot \mathcal{R}\big(\mathcal{G}(\mathcal{F}_0) + \mathcal{H}(\mathcal{F}_1^{\text{ enc}})\big)\Big),
\end{equation}
where $\mathcal{G}$ is a $1 \times 1$ convolution layer followed by a Group Normalization (GN), $\mathcal{H}$ is a $1 \times 1$ convolution followed by a GN and a bilinear upsampling, and $\mathcal{R}$ is a $3 \times 3$ convolution followed by a GN, a ReLU, and a $1 \times 1$ convolution. $\mathcal{F}_0$ and $\mathcal{F}_1^{\text{ enc}}$ represent the per-pixel feature maps produced by the backbone and encoder, respectively.

\vspace{0.5em}\noindent \textbf{Captioning Head.}
Following~\cite{xu2015show}, we employ a classic Long Short-Term Memory (LSTM) network that generates a caption by producing one word at each time step, conditioned on a context vector, the previous hidden state, and the previously generated words.
\begin{equation}
\begin{aligned}
    \begin{pmatrix}
     i_{t} \\
     f_{t}\\
     o_{t}\\
    g_{t}
    \end{pmatrix} &= \begin{pmatrix}
     \sigma \\
     \sigma  \\
      \sigma \\
    \tanh
    \end{pmatrix} T_{D+m+n,n}\begin{pmatrix}
     E_{y_{t-1}}\\
     h_{t-1}\\
    \hat{z_{t}} 
    \end{pmatrix},\\
    c_{t} &= f_{t} \odot c_{t-1} + i_{t} \odot g_{t}, \\
    h_{t} &= o_{t} \odot \tanh(c_{t}).
\end{aligned}
\end{equation}
Here, $i_{t}$, $f_{t}$, $o_{t}$, $g_{t}$, and $h_{t}$ represent the input, forget, memory, output, and hidden states of the LSTM, respectively. The context vector denoted as $\hat{z} \in \mathbb{R}^{D}$, captures the visual information associated with a specific input location. The embedding matrix $E\in \mathbb{R}^{m\times K}$ is also considered. Let $m$ and $n$ represent the embedding and LSTM dimensionality, respectively, while $\sigma$ and $\odot$ denote the logistic sigmoid activation and element-wise multiplication, respectively.

\vspace{0.5em}\noindent \textbf{Trainable Parameters.}
The trained parameters comprise two parts: in the first stage, the parameters of each task head are trained, while in the second stage, the weights of LoRA are trained. In terms of parameter size settings, taking EVA-ViT-E as an example, the total parameter size of all task heads amounts to $36.8$M. We set the size of the two parts to be roughly equal, thus setting the rank of LoRA to $64$, resulting in a parameter size of $29.4$M.

\section{Main Experiments}
\label{exs}
\subsection{Evaluation Benchmarks}
We focus on performance on out-of-domain tasks that are not included as part of the supervised vision finetuning, encompassing both visual and visual-linguistic benchmarks:

(1) Optical Character Recognition (OCR): After freezing the vision transformer and its corresponding LoRA weights, we follow the approach in ~\cite{atienza2021vision} to train a lightweight head for optical character recognition. Utilizing the frozen backbone weights, we employ the MJSynth~\cite{jaderberg2014synthetic} and SynthText~\cite{gupta2016synthetic} datasets for training and evaluate the performance on a combined set of multiple OCR datasets, including IC$03$~\cite{lucas2005icdar}, IC$13$~\cite{karatzas2013icdar}, IC$15$~\cite{karatzas2015icdar}, SVTP~\cite{phan2013recognizing}, SVT~\cite{wang2011end}, and IIIT~\cite{mishra2012scene}. 

(2) Grounded Object Identification: We evaluate the model's performance on the M$^{3}$IT dataset~\cite{li2023m}, which involves classifying an object specified in an image. 

(3) Image Classification: We replace EVA-CLIP's visual encoder with the fine-tuned EVA-ViT and perform zero-shot classification on ImageNet-1K~\cite{deng2009imagenet} and its variants (ImageNet-A~\cite{hendrycks2021natural}, ImageNet-R~\cite{hendrycks2021many}, ImageNet-Sketch~\cite{wang2019learning}). Additionally, we perform few-shot probing on several other datasets~\cite{fei2004learning,maji2013fine,krause20133d,krizhevsky2009learning}.

(4) Image-Text Retrieval: We examine the zero-shot retrieval performance on COCO~\cite{chen2015microsoft} and Flickr30K~\cite{plummer2015flickr30k} for both EVA-CLIP-E and BLIP-2, where the vision encoder is replaced by the fine-tuned EVA-ViT-E and EVA-ViT-G, respectively.

(5) Visual Question Answering: After fine-tuning the visual encoder of BLIP-2, we conduct a quantitative evaluation of the zero-shot visual question answering task on VQAv2~\cite{goyal2017making} and OK-VQA~\cite{marino2019ok}. 

\subsection{Implementation Details}
During the first stage of training, Detr~\cite{carion2020end} serves as the detection head, featuring six encoder layers and six decoder layers. The encoder dimension is $128$, the decoder dimension is $256$, and the MLP dimension is $1024$. For the segmentation head, Mask2former~\cite{cheng2022masked} consists of six encoder layers and nine decoder layers. The encoder dimension is $256$, the encoder MLP dimension is $512$, the decoder dimension is $256$, and the decoder MLP dimension is $1024$. Both Detr and Mask2former share the following settings: the number of attention heads is $8$, the number of input query embeddings is $100$, the batch size is $1$ per GPU, the number of feature levels is $1$, and the learning rate is $5e-5$. Both models are trained for $150$k iterations. 

With respect to the captioning head, we primarily adhere to the settings presented in ~\cite{xu2015show}. The LSTM encoder and decoder dimensions are both $384$, the batch size is $32$ per GPU, the learning rate is $4e-4$, and the training proceeds for $100$k iterations. All task head training utilizes the AdamW optimizer~\cite{loshchilov2017decoupled}, embraces a cosine learning rate strategy, and incorporates a warmup of $2$k iterations. The training for each task head is executed using $8$ Nvidia Volta V$100$-SXM$2$-$32$GB GPUs. The training of various task heads can be conducted concurrently, with the first stage of training requiring less than $2$ days to finish.

During the second stage of training, we jointly train EVA-ViT on multiple tasks. At each iteration, we randomly select a task to fill a batch of samples. We simply assign a comparable sampling probability for each task ($0.4$ for captioning, $0.3$ for both detection and segmentation). In our implementation, we employ $8$ NVIDIA Volta V$100$-SXM$2$-$32$GB GPUs in a distributed manner, using PyTorch~\cite{paszke2019pytorch}. To alleviate the CUDA memory pressure, we have enabled optimizer state sharding. It uses the ZeRO optimizer state sharding method as described in ~\cite{rajbhandari2020zero}. Additionally, gradient checkpointing~\cite{chen2016training} is activated. The AdamW optimizer is utilized with a learning rate of $1e-5$ and a warm-up cosine learning rate schedule (using $2000$ warm-up iterations). 

The training process continues for $50$k iterations, with checkpoints saved every $5$k iterations. The second stage of training requires less than $2$ days to complete. We denote the model after $5$k iterations as the \textbf{default} ViSFT setting, as it shows improvement on the majority of benchmarks. 

\subsection{Main Results}

\noindent \textbf{Optical Character Recognition (OCR).}
Optical Character Recognition (OCR) aims to extract textual information from images, posing a fine-grained and challenging task due to the variability in fonts, colors, sizes, and orientations of the text within images. Consequently, OCR serves as an effective benchmark to evaluate the capability of a vision foundation model in capturing the fine-grained and semantic information of an image. 

In line with the methodology proposed in ~\cite{atienza2021vision}, we implement a vision transformer as the backbone of our model, freezing both the backbone and its corresponding LoRA weights. Following this, we train a 4-layer lightweight transformer head specifically designed for the OCR task. To evaluate the effectiveness of our approach, we perform experiments on a diverse collection of OCR datasets~\cite{lucas2005icdar,karatzas2013icdar,karatzas2015icdar,phan2013recognizing,wang2011end,mishra2012scene} and report the average accuracy. The results presented in Table~\ref{tab:ocr} demonstrate that after applying the ViSFT, the performance of optical character recognition can be improved by at least $2.5$ points, which indicates that the vision transformer effectively regains fine-grained information and is able to capture both the intricate details and semantic information of the image.

\begin{table}[t]
\caption{Evaluation of optical character recognition performance before and after Vision SFT implementation. ``Accuracy'' represents the ratio of correct word instances to the total number of word instances (\%). ``Iters'' refers to the number of iterations updated during the second stage.}
    \label{tab:ocr}
\small
    \centering
\resizebox{0.5\textwidth}{!}{
\begin{tabular}{c|c|c|c}
\toprule
 Model      & Params & Iters & Accuracy   \\
  \midrule
      EVA-ViT-G & 1.0B & 0k  & 44.4   \\
      EVA-ViT-G\textsubscript{ViSFT} & 1.0B & 5k& 46.9(+2.5) \\
      EVA-ViT-G\textsubscript{ViSFT} & 1.0B & 15k& 47.6(+3.2) \\
    \bottomrule
\end{tabular}}
    
    
\end{table}

\vspace{0.5em}\noindent \textbf{Grounded Object Identification.}
Grounded Object Identification (GOI) involves classifying a specified object in an image using the [CLS] token feature of vision transformers. This fine-grained task was not seen during EVA-CLIP's pretraining or our ViSFT. After probing the classification head for 30 epochs on the M$^{3}$IT dataset, both EVA-ViT-G and EVA-ViT-E exhibit an enhancement ranging from $0.3$ to $0.6$ points, as depicted in Table~\ref{tab:goi}. The improvement is more pronounced for EVA-ViT-G, which is a smaller model. These results indicate that ViSFT can bolster the model's generalization performance, with more significant improvements observed in smaller models, which possess fewer parameters and are more prone to losing fine-grained information during image-text pretraining.
\begin{table}[t]
\caption{Performance of grounded object identification under various conditions. We report the Top-1 accuracy (\%) on M$^{3}$IT's validation set with improvements denoted in brackets, e.g., $(+0.6)$.
    ``Iters'' refers to the number of iterations updated during the second stage.} 
    \label{tab:goi}
\small
    \centering
\resizebox{0.5\textwidth}{!}{
\begin{tabular}{c|c|c|c}
\toprule
 \multirow{2}{*}{Model}       & \multirow{2}{*}{Params} & \multirow{2}{*}{Iters} & M$^{3}$IT~\cite{li2023m} val  \\
   & & & Top-1 Acc   \\
  \midrule
      EVA-ViT-G & 1.0B & 0k&52.3   \\
      EVA-ViT-G\textsubscript{ViSFT} & 1.0B & 5k& 52.9(+0.6)  \\
      \midrule 
      EVA-ViT-E  & 4.4B &0k&54.9   \\
      EVA-ViT-E\textsubscript{ViSFT} & 4.4B &5k & 55.2(+0.3) \\
    \bottomrule
\end{tabular}}

\end{table}

\noindent \textbf{Image Classification.} In Table~\ref{tab:cls}, we further exhibit the effectiveness and robustness of our approach across zero-shot and few-shot image classification benchmarks.  We conduct zero-shot classification on EVA-CLIP-E before and after ViSFT, observing improvements across ImageNet-1K and its variant datasets. Compared to EVA-CLIP, which requires the addition of 300M extra parameters and retraining on $144$ A$100$ GPUs to achieve an increase from $81.9\%$ to $82.0\%$ on ImageNet-1K~\cite{sun2023eva}, ViSFT demonstrates its efficiency. Notable enhancements are evident on datasets consisting of adversarial examples, such as ImageNet-A~\cite{hendrycks2021natural} (increasing from $82.1\%$ to $82.4\%$), indicating that fine-grained information can strengthen the model's robustness to real-world perturbations. Furthermore, results from few-shot probing suggest that our proposed method exhibits good generalization capabilities.

\begin{table}
  \centering
  \caption{Zero-shot image classification results on ImageNet-1K and its variants (a). Few-shot probing results on some additional classification datasets (b). Top-1 accuracy (\%) on validation sets is reported. Results exhibiting notable improvements are emphasized in \textbf{Bold}. The number of iterations updated during the second stage in this case is also 5k.}
  \label{tab:cls}
  \begin{subtable}{0.43\textwidth}
    \centering
    \caption{Zero-shot results}
    \resizebox{1.\textwidth}{!}{
    \begin{tabular}{c|cccc}
\toprule
      &  \rotatebox{90}{ImageNet-A~\cite{hendrycks2021natural}} & \rotatebox{90}{ImageNet-R~\cite{hendrycks2021many}}  & \rotatebox{90}{ImageNet-S~\cite{wang2019learning}} & \rotatebox{90}{ImageNet-1K~\cite{deng2009imagenet}} \\
  \midrule
      EVA-CLIP-E~\cite{sun2023eva} &  82.1 & 94.5 & 71.6 & 82.0 \\
      EVA-CLIP-E\textsubscript{ViSFT} & \textbf{82.4} & 94.6 & 71.7 & 82.1   \\
    \bottomrule
    \end{tabular}}
  \end{subtable}
  \hfill
  \begin{subtable}{0.43\textwidth}
    \centering
    \caption{Few-shot results}
    \resizebox{1.\textwidth}{!}{
    \begin{tabular}{c|cccc}
\toprule
      &  \rotatebox{90}{Caltech101~\cite{fei2004learning}} & \rotatebox{90}{Aircraft~\cite{maji2013fine}}  & \rotatebox{90}{Stanford Cars~\cite{krause20133d}} & \rotatebox{90}{CIFAR-100~\cite{krizhevsky2009learning}} \\
  \midrule
      EVA-CLIP-E~\cite{sun2023eva} &  92.4 & 68.1 & 91.6 & 90.2 \\
      EVA-CLIP-E\textsubscript{ViSFT} & \textbf{94.3} & \textbf{69.7} & \textbf{92.5} & \textbf{90.9}  \\
    \bottomrule
    \end{tabular}}
  \end{subtable}
\end{table}

\vspace{0.5em} \noindent \textbf{Image-Text Retrieval.}
Table~\ref{tab:ret} presents the zero-shot image and text retrieval results on Flickr30K and COCO. Upon implementing ViSFT, BLIP-2 exhibits enhancements in both text and image retrieval, with a bit more impact observed in image retrieval tasks. This is attributable to the model is able to better understand and extract relevant features from images when paired with corresponding texts. Due to resource limitations, we opted not to retrain BLIP-2's q-former from scratch. Instead, we utilized the pre-trained weights of the q-former and fine-tuned it for 1k iterations on VG~\cite{krishna2017visual} and COCO Caption datasets using the obtained LoRA weights. These datasets represent a small fraction of the larger BLIP-2 pretraining dataset.

We further conducted evaluations on EVA-CLIP, due to the limited resources, we did not fine-tune the text encoder of EVA-CLIP utilizing LoRA weights obtained through ViSFT. Instead, we reported the results of another checkpoint at different update iterations (e.g. $50$k). we observed phenomena similar to those of BLIP-2, which further substantiates our conclusions.

\begin{table} [t]
    \centering
    \caption{Comparison of image-text retrieval performance across various settings. Results are assessed using Recall@5 (\%). Performance for both Flickr30K and COCO datasets are reported, with evaluations conducted on EVA-CLIP and BLIP-2. ``Iters'' refers to the number of iterations updated during the second stage.}
    \label{tab:ret}
\resizebox{0.75\textwidth}{!}{
    \begin{tabular}{c|c|cc|cc}
    \toprule
 \multirow{3}{*}{Model} & \multirow{3}{*}{Iters}  & \multicolumn{2}{c|}{Text Retrieval} & \multicolumn{2}{c}{ Image Retrieval} \\
  & &  Flickr30k  &   COCO &  Flickr30k  &   COCO  \\
  & &  R@5  &   R@5 &  R@5 &   R@5  \\
   \midrule
   BLIP-2 ViT-G~\cite{li2023blip} & 0k& 99.9  & 94.2  & 96.8  & 84.0\\
      BLIP-2 ViT-G\textsubscript{ViSFT} & 5k  & 99.9  & 95.1(+0.9)  & 97.3(\textbf{+0.5})  & 85.1(\textbf{+1.1})\\
    \midrule
      EVA-CLIP-E~\cite{sun2023eva}  & 0k  & 99.4 & 87.6  & 94.3  & 74.9   \\
    EVA-CLIP-E\textsubscript{ViSFT} & 5k  & 99.4 &  87.7(+0.1)   & 94.3  & 75.2(+0.3)   \\
    EVA-CLIP-E\textsubscript{ViSFT}  & 50k  & 99.5(+0.1) &  87.6  & 94.8(\textbf{+0.5})  & 76.0(\textbf{+1.1})   \\
   \bottomrule   
    \end{tabular}

}
    
\end{table}

\vspace{0.5em}\noindent \textbf{Visual Question Answering.}
We assessed the zero-shot visual question answering performance of BLIP-2 ViT-G OPT using benchmarks such as VQAv2~\cite{goyal2017making} and OK-VQA~\cite{marino2019ok}. As mentioned before, we fine-tuned the pre-trained q-former for 1k iterations on VG and COCO Caption using LoRA weights obtained through ViSFT. Table~\ref{tab:vqa} demonstrates the effectiveness of our approach on both benchmarks. The improvement is a bit more pronounced on OK-VQA, suggesting that ViSFT provides benefits for out-of-domain datasets.

\begin{table}[t]
\small
    \centering
    \caption{Zero-shot visual question answering results. Metrics include accuracy for VQAv2, and OK-VQA(\%). Evaluations are conducted on BLIP-2 ViT-G OPT$_{2.7B}$ (designated as OPT$^{s}$).}
    \label{tab:vqa}
\resizebox{0.7\textwidth}{!}{
\begin{tabular}{c|c|c|cc}
\toprule
 Model      & Params  & Iters   
   & VQAv2 & OK-VQA   \\
   \midrule
   BLIP-2 ViT-G OPT$^{s}$~\cite{li2023blip}  & 3.8B & 0k& 51.9 & 31.5 \\
   BLIP-2 ViT-G OPT$^{s}$\textsubscript{ViSFT}  & 3.8B & 5k& \textbf{53.0} & \textbf{32.8}  \\
    \bottomrule
\end{tabular}}
    
    
\end{table}

\subsection{Ablation Studies}
\label{ablation}
In the subsequent sections, we examine the critical designs of our ViSFT in conjunction with EVA-CLIP-E. Unless explicitly stated, image-text retrieval performance on the COCO dataset is evaluated. All ablation studies are conducted on the model after 5k iterations of updates, which is the default setting for ViSFT.

\vspace{0.5em}\noindent \textbf{Effects of LoRA Rank.}
In the rank configuration for LoRA, as mentioned before, we employed the default value of $r = 64$, which results in comparable parameter sizes for LoRA and task heads within our experimental setup. Table~\ref{tab:rank} demonstrates that LoRA exhibits competitive performance when $r >= 32$. Consequently, we maintain the original default configuration, and the additional costs incurred compared to smaller rank settings are negligible.

\begin{table}[t]
    \centering
    \caption{Ablation analysis of  Ablation analysis of LoRA with varying ranks (a). And training data size (b): K$\%$ indicates the use of K$\%$ of the available training data. Results are presented for text retrieval (R@$5$), and image retrieval (R@$5$).}
    \label{tab:rank}
  \begin{subtable}{0.4\textwidth}
    \centering
    \caption{Rank of LoRA}
    \label{tab:rank}
    \resizebox{1.\textwidth}{!}{
   \begin{tabular}{c|cc}
\toprule
    \multirow{3}{*}{Rank}    & \multicolumn{2}{c}{COCO}  \\
     & Text Retrieval  & Image Retrieval  \\
     & R@5 & R@5 \\
   \midrule
   $r = 16$  & 87.7 & 75.1 \\
   $r = 32$  & 87.8 & 75.1 \\
    $r = 64$  & 87.7 & 75.2 \\
    \bottomrule
\end{tabular}}
  \end{subtable}
  \hfill
  \begin{subtable}{0.43\textwidth}
    \centering
    \caption{Training data size}
    \label{tab:datasize}
    
    \resizebox{1.\textwidth}{!}{
 \begin{tabular}{c|cc}
\toprule
    \multirow{3}{*}{Data Size}  & \multicolumn{2}{c}{COCO}  \\
     & Text Retrieval  & Image Retrieval  \\
     & R@5 & R@5 \\
   \midrule
   $25\%$  & 87.6 & 75.1 \\
   $50\%$  & 87.6  & 75.1 \\
    $100\%$  & 87.7 & 75.2 \\
    \bottomrule
    \end{tabular}}
  \end{subtable}
  \end{table}

\vspace{0.5em}\noindent \textbf{Training Data Size.}
Table~\ref{tab:datasize} demonstrates that using the full training dataset results in marginally better performance, implying that there could be potential for improvement if more data with annotations similar to COCO is leveraged. We postpone this investigation to future work, as the current impact of training data size is minimal. For example, increasing the training data size from $25\%$ to $100\%$ only leads to a $0.1\%$ enhancement in the image-text retrieval task.

\begin{table}[t]
    \centering
    \caption{Ablation analysis of training strategies. Results are presented for zero-shot image classification and image-text Retrieval.}
    \label{tab:headlr}
\resizebox{0.6\textwidth}{!}{
  \begin{tabular}{c|c|cc}
    \toprule
 \multirow{3}{*}{Strategies} & ImageNet-1K  & \multicolumn{2}{c}{COCO}  \\
     & Classification & Text Retrieval & Image Retrieval \\
     & Top-1 & R@5 & R@5\\
     \midrule
     Two-stage & 82.1 & 87.7 & 75.2 \\ 
    One-stage  & 82.0~$\downarrow$ & 87.7 & 75.0~$\downarrow$ \\ 
    \bottomrule
  \end{tabular}}
    
    
\end{table}

\vspace{0.5em} \noindent \textbf{Training Strategies.}
There are two potential strategies for performing vision fine-tuning. The classic multi-task approach entails simultaneous fine-tuning of both the task heads and the backbone in a single-stage training process. However, as Table~\ref{tab:headlr} demonstrates, this method yields suboptimal performance. As previously mentioned, fine-grained information learned from different annotations can be trapped within the task heads. Therefore, we propose the two-stage method and the results indicate that this strategy performs better.

\vspace{0.5em}\noindent \textbf{Selection of Task Types.}
In our default configuration, we adopt object detection, image captioning, and instance segmentation on COCO. To analyze the effects of various tasks, we conduct experiments by either adding new tasks, such as pose estimation, replacing instance segmentation with panoptic segmentation, or independently removing each task from the joint-training tasks. For pose estimation, we employ the ViTPose task head, which utilizes a vision transformer as the backbone and requires only a single-scale input feature. For panoptic segmentation, which combines instance segmentation and semantic segmentation, we maintain the use of the mask2former head to ensure a fair comparison.

Table~\ref{tab:task_type} demonstrates that adding a new task, such as pose estimation, does not yield further performance improvements. This is reasonable, as not all images in COCO contain person instances that would benefit from pose keypoint annotations. A similar phenomenon can be observed in instruction tuning~\cite{wei2021finetuned}: not all task clusters benefit the foundation model.

The results for instance segmentation and panoptic segmentation are competitive, as semantic annotations are more coarse-grained than instance annotations. This indicates that instance annotations possess sufficient granularity for effectively performing our ViSFT.

Upon the removal of any of the three tasks, the model starts to over-optimize for a specific task, such as image-text retrieval, resulting in a decline in the zero-shot image classification performance. This aligns with observations from instruction tuning~\cite{wei2021finetuned}, emphasizing the importance of task diversity for executing supervised fine-tuning.

\begin{table}[t]
  \centering
  \caption{Ablation analysis of task type selection. The evaluation focuses on zero-shot image classification and image-text retrieval. Default setting incorporates object detection, instance segmentation and image captioning. ``\textsubscript{w/}" denotes ``with", ``\textsubscript{w/o}" signifies ``without" and ``\textsubscript{r/ panoptic}" represents ``instance segmentation is replaced by panoptic segmentation".}
  \label{tab:task_type}
  \resizebox{0.7\linewidth}{!}{
  \begin{tabular}{c|c|cc}
    \toprule
 \multirow{3}{*}{Setting} & ImageNet-1K  & \multicolumn{2}{c}{COCO}  \\
     & Classification & Text Retrieval & Image Retrieval \\
     & Top-1 & R@5 & R@5\\
     \midrule
     Default & 82.1 & 87.7 & 75.2 \\ 
    \textsubscript{w/ pose}  & 82.1 & 87.7 & 75.1~$\downarrow$ \\ 
    \textsubscript{r/ panoptic}&   82.1 & 87.8~$\uparrow$ & 75.1~$\downarrow$ \\ 
    \midrule
    \textsubscript{w/o detection}&  82.0~$\downarrow$ & 87.9~$\uparrow$ & 75.2 \\ 
    \textsubscript{w/o segmentation}&  82.0~$\downarrow$ & 87.8~$\uparrow$ & 75.1~$\downarrow$ \\ 
    \textsubscript{w/o caption}&  82.0~$\downarrow$ & 87.8~$\uparrow$ & 75.2 \\ 
    \bottomrule
  \end{tabular}
  }
  
\end{table}

\subsection{Visualization}

\begin{figure}
\centering
\begin{subfigure}{0.45\textwidth}
    \includegraphics[width=\textwidth]{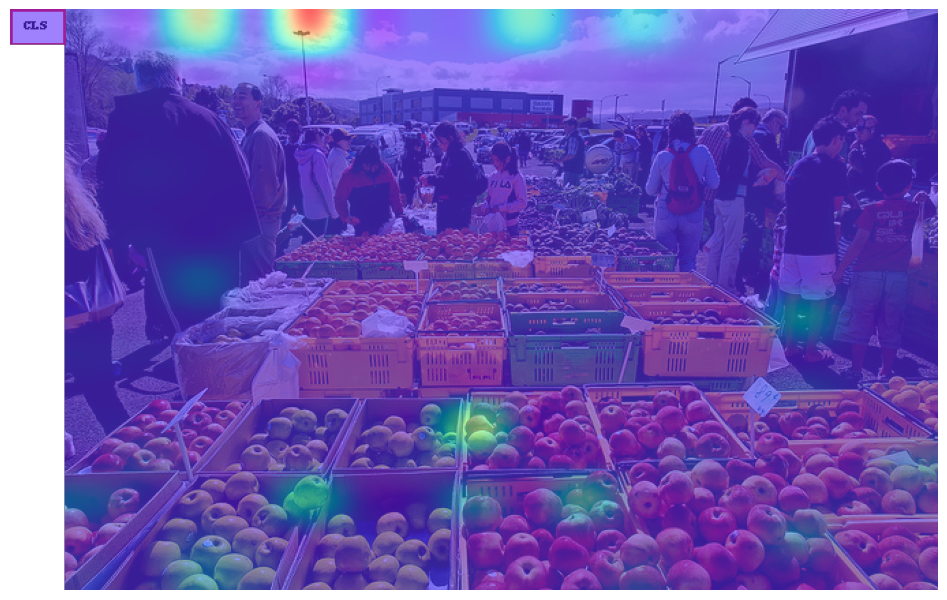}
    \caption{$w/o$ ViSFT}
    \label{fig:first}
\end{subfigure}
\hfill
\begin{subfigure}{0.45\textwidth}
    \includegraphics[width=\textwidth]{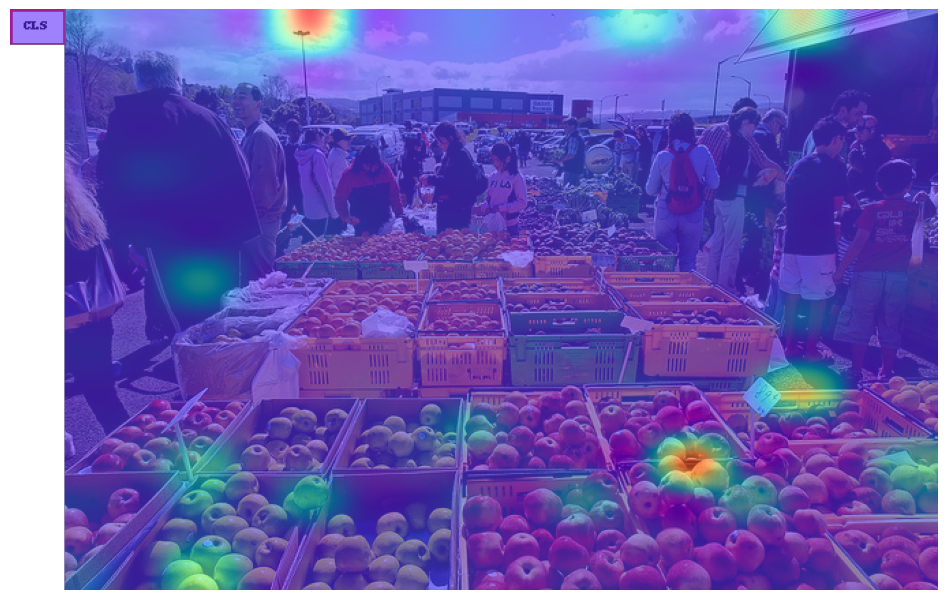}
    \caption{$w/$ ViSFT}
    \label{fig:second}
\end{subfigure}
\caption{Visualization of [CLS] token's attention distribution. Experiments are conducted on the last layer of EVA-ViT-G. Attended image patches are highlighted.}
\label{fig:vis}

\end{figure}

To further substantiate the efficacy of our approach, we have conducted a visualization of ViSFT. The image patches of EVA-ViT-G are reshaped into a 2D configuration following the insertion of the [CLS] token, and we visualize the attention distribution of the [CLS] token across these patches. As depicted in Figure~\ref{fig:vis}, after applying ViSFT, the [CLS] token not only attends to nearby patches (highlighted at the top of the images) but also focuses on more distant objects. This suggests that ViSFT assists vision foundation models in capturing fine-grained information from image patches.

\section{Conclusion}

Drawing inspiration from natural language processing, we explore the potential of fine-grained supervised fine-tuning (SFT) to enhance the generalization and representation capabilities of vision foundation models after pretraining. We propose a two-stage method, termed "ViSFT" to effectively unleash the fine-grained knowledge embedded within these models. Through our lightweight training process, the performance of vision foundation models exhibits improvements across a wide range of out-of-domain benchmarks in both visual and vision-linguistic scenarios.

\bibliographystyle{splncs04}
\bibliography{main}
\clearpage  
\appendix
\clearpage
\setcounter{page}{1}

\section{ViSFT Procedure}
The ViSFT training process can be described in Algorithm~\ref{alg:s1} and Algorithm~\ref{alg:s2}, which obtain compatible in-domain task head  $T_{n}^{*}$ and learned LoRA weights $\Delta W^{*}$, respectively.

Upon acquiring the learned LoRA weights $\Delta W^{*}$, evaluations on out-of-domain benchmarks can be outlined in Algorithm~\ref{alg:eval}.

\begin{algorithm}
	\caption{Stage1 Training} 
	\begin{algorithmic}[1]
            \Require Training dataset $D(\mathbf{x},\mathbf{y})$; Pretrained vision foundation model $M$
            \State Initialize an in-domain task head $T_{n}$, for $n \in \{1,\ldots, N\}$ and freeze $M$
		\For {$i=1,2,\ldots$} \Comment{Can be executed in parallel }
                    \State Extract feature $\textbf{f} = M(\mathbf{x})$ for input $\mathbf{x}$ 
				\State Minimize $L_{n}(\mathbf{y}, T_{n}(\mathbf{f}))$ on $D$ to obtain $T_{n}^{*}$
		\EndFor
	\end{algorithmic} 
 \label{alg:s1}
\end{algorithm}

\begin{algorithm}
	\caption{Stage2 Training}
	\begin{algorithmic}[1]
		\Require In-domain task head $T_{n}^{*}$; Pretrained vision foundation model $M$; Sampling
 probability $\alpha_n$, $n \in \{1, \ldots, N\}$
    \State Initialize LoRA weights $\Delta W$, freeze $M$ and $T_{n}^{*}$, $n \in \{1, \ldots, N\}$
		\For {$i=1,2,\ldots$} 
                    \State Select an in-domain task $T_{n}^{*}$ according to P($\alpha_n$)
                    \State Extract feature $\mathbf{f}^{'} = M(\mathbf{x}; \Delta W)$ for input $\mathbf{x}$ 
				\State Minimize $L_{n}^{'}(\mathbf{y}, T_{n}(\mathbf{f^{'}}))$ on $D$ to obtain $\Delta W^{*}$
		\EndFor
	\end{algorithmic} 
 \label{alg:s2}
\end{algorithm} 

\begin{algorithm}
	\caption{Evaluation}
	\begin{algorithmic}[1]
		\Require Pretrained vision foundation model $M$; Learned LoRA weights $\Delta W^{*}$; Out-of-domain benchmark $T_{o}$; Evaluation dataset $E_{o}(\mathbf{x},\mathbf{y})$, $o \in \{1, \ldots, O\}$
            \State Initialize results list $R_{o}$
		\For {$\mathbf{x}$ in $E_{o}(\mathbf{x})$} 
                    \State Extract feature $\mathbf{f}^{*} = M(\mathbf{x}; \Delta W^{*})$ for input $\mathbf{x}$ 
				\State Predicting $R_{o} = [R_{o}, T_{o}(\mathbf{f^{*}})$]
		\EndFor
  \State Accumulate results: $Metric(E_{o}(\mathbf{y}),R_{o})$ on $E_{o}$ 
	\end{algorithmic} 
 \label{alg:eval}
\end{algorithm} 


\section{Licenses of Datasets}
\noindent\textbf{ImageNet-1K} \cite{deng2009imagenet} is subject to the ImageNet terms of use \citeappendix{imagenetterms}.

\vspace{0.5em}\noindent  \textbf{ImageNet-A}~\cite{hendrycks2021natural} is subject to the ImageNet-A terms of use~\citeappendix{imageAlicense}.

\vspace{0.5em}\noindent  \textbf{ImageNet-R}~\cite{hendrycks2021many} is subject to the ImageNet-R terms of use~\citeappendix{imageRlicense}.

\vspace{0.5em}\noindent  \textbf{ImageNet-Sketch}~\cite{wang2019learning} is subject to the ImageNet-Sketch terms of use~\citeappendix{imageSlicense}.


\vspace{0.5em}\noindent  \textbf{Caltech-101}~\cite{fei2004learning} is subject to the Caltech-101 terms of use~\citeappendix{caltechlicense}.

\vspace{0.5em}\noindent  \textbf{Aircraft}~\cite{maji2013fine} is subject to the FGVC-Aircraft terms of use~\citeappendix{aircraftlicense}.

\vspace{0.5em}\noindent  \textbf{IC03}~\cite{lucas2005icdar} is subject to the ICDAR 2003 terms of use~\citeappendix{ic03license}.

\vspace{0.5em}\noindent  \textbf{IIIT}~\cite{mishra2012scene} is subject to the IIIT5k-word terms of use~\citeappendix{iiitlicense}.

\vspace{0.5em}\noindent  \textbf{MJSynth}~\cite{jaderberg2014synthetic} is subject to the MJSynth terms of use~\citeappendix{mjlicense}.

\vspace{0.5em}\noindent  \textbf{SynthText}~\cite{gupta2016synthetic}is subject to the SynthText terms of use~\citeappendix{stlicense}.

\vspace{0.5em}\noindent  \textbf{M$^{3}$IT}~\cite{li2023m} is subject to the M$^{3}$IT terms of use~\citeappendix{mitlicense}.

\vspace{0.5em}\noindent  \textbf{COCO}~\cite{lin2014microsoft} is subject to the COCO terms of use~\citeappendix{cocoterms}.

\vspace{0.5em} \noindent \textbf{Visual Genome}~\cite{krishna2017visual} is licensed under a Creative Commons Attribution 4.0 International License \citeappendix{vgterms}.

\vspace{0.5em}\noindent  \textbf{Flickr30K}~\cite{plummer2015flickr30k} is subject to the Flickr terms of use~\citeappendix{flickr2020terms}.

\vspace{0.5em}\noindent  \textbf{VQAv2}~\cite{goyal2017making} is subject to the VQAv2 terms of use~\citeappendix{vqalicense}.

\vspace{0.5em}\noindent  \textbf{OK-VQA}~\cite{marino2019ok} is subject to the OK-VQA terms of use~\citeappendix{okvqalicense}.

\renewcommand{\refname}{Appendix References}
\vspace{2em}
\bibliographystyleappendix{splncs04}
\bibliographyappendix{main}

%
%

\end{document}